\documentclass[11pt,a4paper]{article}
\usepackage[hyperref]{eacl2021}
\usepackage{times}
\usepackage{latexsym}

\usepackage{caption}
\usepackage{subcaption}
\usepackage{tikz}
\usetikzlibrary{shapes,arrows}
\usetikzlibrary{positioning}
\usetikzlibrary{calc}
\usetikzlibrary{fit}
\usetikzlibrary{shapes,decorations,arrows,calc,arrows.meta,fit,positioning}
\definecolor{goodbluebar}{RGB}{114, 147, 203}
\definecolor{goodorangebar}{RGB}{225, 151, 76}
\definecolor{goodgreenbar}{RGB}{132, 186, 91}
\definecolor{goodredbar}{RGB}{211, 94, 96}
\definecolor{goodblackbar}{RGB}{128, 133, 133}
\definecolor{goodpurplebar}{RGB}{144, 103, 167}
\definecolor{goodwinebar}{RGB}{171, 104, 87}
\definecolor{goodgoldbar}{RGB}{204, 194, 16}

\definecolor{goodblue}{RGB}{57, 106, 177}
\definecolor{goodorange}{RGB}{218, 124, 48}
\definecolor{goodgreen}{RGB}{62, 150, 81}
\definecolor{goodred}{RGB}{204, 37, 41}
\definecolor{goodblack}{RGB}{83, 81, 84}
\definecolor{goodpurple}{RGB}{107, 76, 154}
\definecolor{goodwine}{RGB}{146, 36, 40}
\definecolor{goodgold}{RGB}{148, 139, 61}

\usepackage{microtype}

\aclfinalcopy 

\usepackage{amsmath}
\usepackage{amssymb}
\usepackage{mathabx}

\title{Ranking Creative Language Characteristics in Small Data Scenarios}
\author{
Julia Siekiera$^{1}$,
Marius Köppel$^{2}$,
Edwin Simpson$^{3,4}$,
Kevin Stowe$^{4}$,
Iryna Gurevych$^{4}$,
Stefan Kramer$^{1}$\\
  \\
  $^1$Dept. of Computer Science and $^2$Institute for Nuclear Physics, Johannes Gutenberg-Universität Mainz,\\
  \texttt{\{siekiera,mkoeppel\}@uni-mainz.de},
  \texttt{kramer@informatik.uni-mainz.de},\\
  $^3$Dept. of Computer Science, University of Bristol, \texttt{edwin.simpson@bris.ac.uk},\\
  $^4$Ubiquitous Knowledge Processing Lab,
  Technische Universität Darmstadt\\
  {\sf https://www.informatik.tu-darmstadt.de/ukp}\\
}

\date{}

\begin{document}
\maketitle
\begin{abstract}
The ability to rank creative natural language provides an important general tool for downstream language understanding and generation. 
However, current deep ranking models require substantial 
amounts of labeled data that are difficult and expensive to obtain for different domains, languages and creative characteristics.
A recent neural approach, the DirectRanker,
promises to reduce
the amount of training data needed but
its application to text isn't fully explored.
We therefore adapt the DirectRanker to provide a new deep model for ranking creative language with small data.
We compare DirectRanker with a Bayesian approach, Gaussian process preference learning (GPPL), which has previously been shown to work well with sparse data. 
Our experiments with sparse training data 
show that 
while the performance of standard neural ranking approaches collapses with small training datasets,
DirectRanker remains effective.
We find that
combining DirectRanker with
GPPL increases performance across different settings by leveraging the complementary benefits of both models.
Our combined approach 
outperforms the previous state-of-the-art on humor and metaphor novelty tasks, increasing Spearman's $\rho$ by 14\% and 16\% on average.
\end{abstract}

\section{Introduction}
Everyday language is characterised by numerous creative and figurative expressions, such as metaphors, jokes or witticisms.
To interpret language correctly, it is important for natural language processing models to recognise such devices, 
so that the system does not interpret a joke or metaphor at literally and can instead identify its intended meaning. 
However, the simple recognition of creative language as a binary classification is usually not sufficient. 
Creative language like humour and metaphor can be present to different degrees, which require different kinds of processing and different responses from conversational agents.
This calls for models that can assess the degree to which a creative quality is present in a piece of text.

If annotators have to assign scores to individual documents, inconsistencies can arise both between annotators and over time across the labels of a single annotator.
Pairwise comparisons between documents simplify the annotation process as annotators only have to solve one decision without needing to calibrate their scores.
The entire ranking can then either be derived by comparing all pairs or at least be estimated from a sparse subset.
Considering the expensive and time consuming annotation process, normally only a small subset of the pairwise comparisons can be obtained on large datasets, requiring the need for an effective estimator for sparse data.

A simple baseline method
is the MaxDiff approach applied to best-worst scaling (BWS) annotations~\cite{marley2005some},
which estimates scores by applying a simple function to counts of the numbers of times that each document was chosen in a preference pair.
However, this method does not consider the text itself, and thus performs poorly when the comparisons are sparse.
We therefore turn to ranking models that make use of the text itself when estimating the score for each instance.

\emph{Gaussian process preference learning (GPPL)} was recently shown to 
   outperform MaxDiff 
   at ranking creative language with sparse crowdsourced data by making use of word embeddings and linguistic features~\citep{simpson}. 
GPPL is a Bayesian approach which explicitly quantifies the uncertainty in the model to provide confidence estimates and reduce overfitting to small and noisy labels.
However, it is a shallow model that applies a pre-determined kernel function to feature vectors representing each instance. Therefore, it requires another method to first map each piece of text to a suitable representation, such as a mean word embedding (MWE)~\cite{mikolov2013distributed}.

Neural network architectures such as RankNet~\cite{rank_net} allow representation learning from pairwise comparisons.
A recent method, ~\emph{DirectRanker}~\cite{koeppel} 
improves label efficiency for document ranking
by generalizing the RankNet model, making it
more suitable for low resource settings.
DirectRanker's architecture fulfills the requirements of a total quasiorder,
which results in faster convergence than other neural network ranking approaches,
as this order does not have to be additionally learned.
This characteristic can be especially useful on small datasets. 
We propose to adapt DirectRanker to creative language.
Our experiments show that depending on the task, DirectRanker is able to outperform the GPPL state-of-the-art on small creative language datasets.

Error analysis for both DirectRanker and GPPL combined with two sentence embedding methods indicates 
differences in the types of document that are misranked by each.
This motivates combining the models through~\emph{stacking},
which we show is able to improve the performance of the two sub-models.

The main contributions of this paper are the following:
\begin{itemize}
    \item Adaptation and evaluation of DirectRanker for the ranking of  creative language, demonstrating competitive ranking quality to state-of-the-art in small data scenarios.
    \item Empirical results showing that combining Bayesian and neural approaches using stacking can address some of the limitations of each method.
    \item An investigation of different text representations, showing a clear benefit to sentence transformer embeddings for the individual models and a combined approach to improve stacking.
\end{itemize}
We will release our complete experimental software on publication.

\section{Related Work}
The general aim of the ranking task is to sort a list of $n$ documents according to a relevance for a given query. Algorithms solving the ranking problem can be divided into three categories.
First, pointwise rankers assign a score to each query-document pair~\citep{probabilistic_regression,poly_retrieval,mcrank}.
This approach is equivalent to document classification.
Second, pairwise models predict which document is more relevant out of two for a given query~\citep{rank_net,gradient_boosting}.
Third, listwise algorithms take a list of documents under consideration to evaluate the training cost~\citep{pairwise_to_listwise}.

Previous research on document ranking combined BERT~\citep{DBLP:journals/corr/abs-1810-04805} with different learning-to-rank methods of all three categories. While~\citet{han2020learning} and~\citet{qiao2019understanding} embed concatenated queries and documents with BERT and fine tune ranking performance by the use of an arbitrary artificial neural network ranker,~\citet{guo2020detext} encode queries and documents independently and apply a two pass ranking model to obtain an online and offline ranking model for industry use cases.~\citet{nogueira2019multi} introduce a multi stage pipeline containing a pointwise and a pairwise BERT ranker to trade off ranking quality against latency.
However, these approaches are evaluated neither for small training data scenarios nor on the difficult task of creative language.
To the best of our knowledge they do not fulfil the requirements of a total quasiorder. 
The DirectRanker should therefore be more suitable for ranking scenarios with sparse datasets.
\section{Methods}\label{sec:methods}
\subsection{GPPL}
GPPL has previously been shown to be an effective solution for ranking
instances of creative language
according to their humorousness and metaphor novelty~\citep{simpson}, as well as for ranking arguments by convincingness, where it outperformed SVM and BiLSTM regression models~\citep{simpsonfinding}. 
GPPL is a Bayesian approach introduced by \citet{10.1145/1102351.1102369}, 
that incorporates a \emph{random utility model} of 
preference~\citep{thurstone1927law,mosteller1951remarks}. Random utility models assume that each instance $x_i$ has a latent \emph{utility}
$u_i$ that represents its value in terms of the compared characteristic. For example, if documents are ranked according to how funny they are, the utility represents their humorousness. In GPPL, pairwise preference labels are then drawn randomly with a probability that depends on the difference between the utilities of the items being compared:
\begin{flalign}
    p(x_1 \succ x_2) = \Phi\left( \frac{u_1 - u_2}{\sqrt 2 \sigma^2} \right),
\end{flalign}
where $x_1 \succ x_2$ indicates that instance $x_1$ was labeled as preferred to $x_2$, $\Phi$ is the probit function, and $\sigma^2$ is a variance parameter. A popular alternative used by other models is the Bradley-Terry model~\citep{bradley1952rank,luce1959possible,plackett1975analysis}. 
GPPL combines the random utility model with a Gaussian process that models the relationship between input features and utilities by learning the function $u_1 = f(x_1)$. This function models the relationships between $f(x_1)$ for different instances with a kernel. Hence, GPPL is able to infer the scores of unseen data not only from the pairwise labels, but also from the additional knowledge of the instance features~\citep{3569}. 

In contrast to artificial neural networks that first optimize model parameters according to maximum likelihood and then make predictions given the chosen parameter values,
Bayesian methods like GPPL integrate over (or marginalize) possible parameter values to compute posterior distributions for test instances that can be used as predictions, meaning that GPPL accounts for confidence in the model itself. For instance, it reduces its confidence when making predictions about test data whose features are very different from the training data. This treatment of model uncertainty can be valuable when faced with sparse and noisy data.
Gaussian processes are also able to represent nonlinear functions with a complexity depending on the amount of training data.
Contradictory annotations can be interpreted as noise, making GPPL suitable for crowdsourced data that contains multiple annotations per instance from non-expert workers. 

The original GPPL implementation has complexity in $\mathcal{O}(n^3)$, which prevents its use for more than a few hundred instances. \citet{simpson2020scalable} therefore employ a scalable version that adapts the
stochastic variational inference method of \citet{pmlr-v38-hensman15} and \citet{JMLR:v14:hoffman13a} to preference learning.
\subsection{DirectRanker}\label{sec:direct_ranker}
\begin{figure}[ht]
	\centering
	 \includegraphics[width=0.49\textwidth]{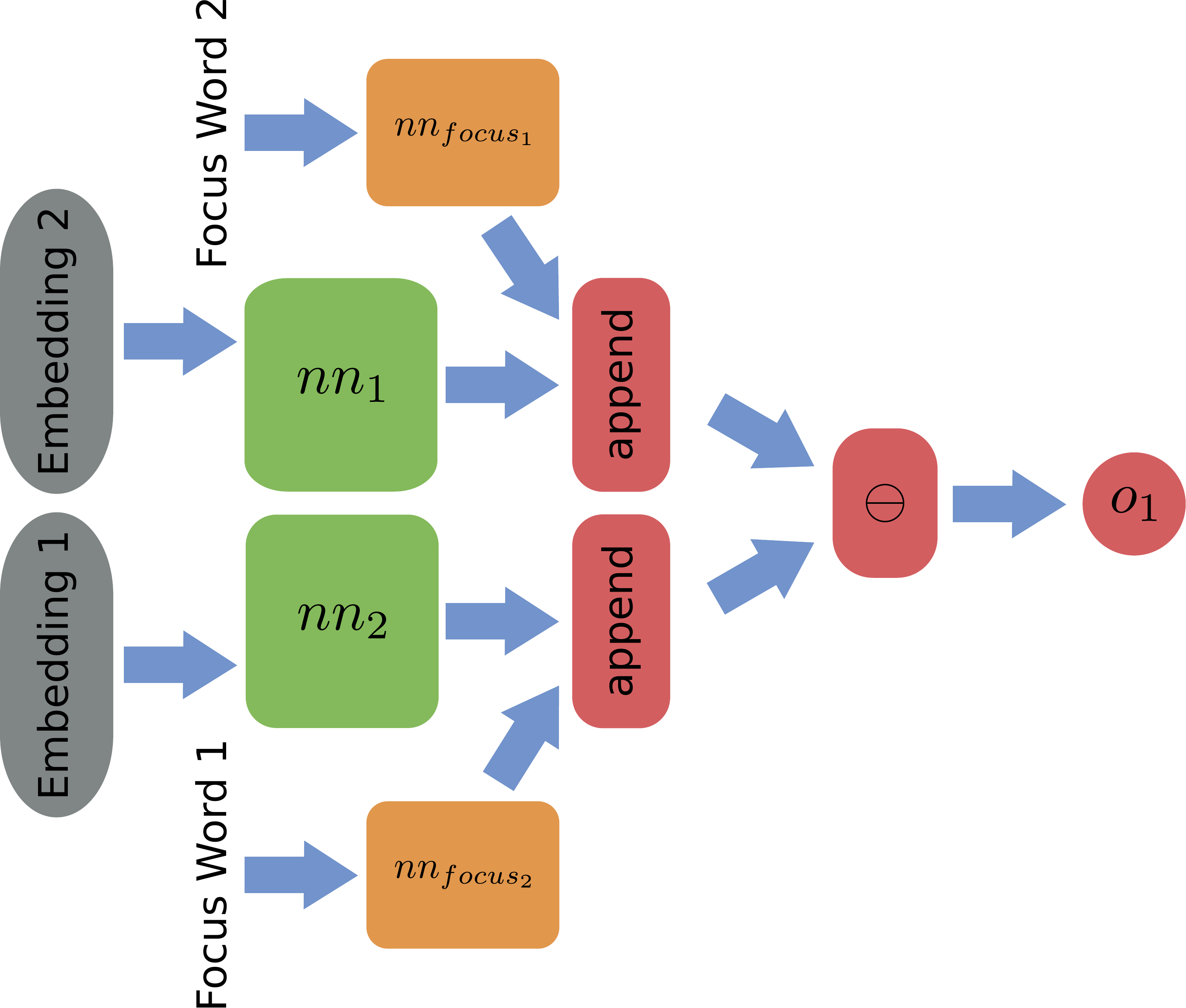}
	\caption{Schematic overview of the adapted DirectRanker architecture. Document embeddings are fed into the parameter sharing networks $nn_1$ and $nn_2$ to generate low-dimensional latent utilities. 
	For datasets containing focus word information, two additional parameter sharing networks $nn_{focus_1}$ and $nn_{focus_2}$ are added. 
	The (appended) two utilities
	are subtracted and fed into the ranking part with output neuron $o_{1}$ that has no bias and uses $\tanh$ as activation.
}
	\label{fig:direct_ranker}
\end{figure}
The DirectRanker~\cite{koeppel} shown in Figure~\ref{fig:direct_ranker} has been introduced as a generalization of the RankNet architecture~\cite{rank_net} to solve the learning-to-rank problem by pairwise document comparisons. The generalization is achieved by implementing a total quasiorder into the model architecture while RankNet needs to learn this order during training.
\citet{koeppel} showed that the DirectRanker is able to outperform numerous state-of-the-art 
pairwise and listwise methods while having a shorter training time.\\
Generally, the model's architecture consists of two sub-models.
The first sub-model is called the feature part, which learns a meaningful low-dimensional latent utility of the input documents.
The second sub-model is called the ranking part and receives the difference of the compressed input tuple and learns the pairwise ranking.
To solve the learning-to-rank problem the model's architecture includes the ranking function:
\begin{equation}
\label{eq:pairwise_ranking}
    o_1 = \tau\left(w\left( \frac{u_1 - u_1}{2} \right) \right),
\end{equation}
where $u_1 = nn_1(x_1)$ and $u_1 = nn_2(x_2)$ are the compression functions for the input feature vectors $x_1$ and $x_2$, $w$ represents the multilayer perceptron (MLP) ranking weights for the last neuron, and $\tau$ is a antisymmetric sign conserving activation.
The relation induced by Equation \ref{eq:pairwise_ranking} is reflexive, total, and transitive by construction enabling a total quasiorder on feature space. The loss function of the relative instance ranking $(x_1,y_1)$ and $(x_2,y_2)$ with $\Delta y=y_1-y_2$ is calculated as in the original DirectRanker paper, namely:
\begin{equation}\label{equ:loss}
L_{\text{rank}}(\Delta y,x_1,x_2)=(\Delta y-o_1(x_1,x_2))^2.
\end{equation}
Reflexivity is enforced since the two networks $nn_1$ and $nn_2$ share their parameters and architecture.
In general, any kind of function approximators can be integrated in the feature part as long as they map the same input to the same output.
Since the output neuron $o_1$ has an antisymmetric sign-conserving neural activation and no bias, the network is guaranteed to produce a total quasiorder~\cite{neural_net_object_ranking}.
The last requirement of the total quasiorder is transitivity that was also proved by~\citet{koeppel}.
Compared to different pairwise~\cite{rank_net} and listwise~\cite{lambdaMart,pairwise_to_listwise,Adarank} ranking models the DirectRanker achieves in only a few training iterations comparable or even better results on both artificial and real world data.
Since the ranking function fulfills the requirements of a total quasiorder, only document pairs where one is more relevant than the other are required as input resulting in a significantly reduced training time.
We decided to further investigate the DirectRanker, since the potential in information retrieval tasks was shown by outperforming the current state-of-the-art models.
In addition to the model's reduced training time on real data, the ranking quality was shown to reach competitive results even on a limited amount of synthetic data~\cite{koeppel}.
Therefore, we believe that the DirectRanker could also be applied in scenarios with sparse pairwise comparisons on creative language.
\subsection{DirectRanker for Text Ranking}
To adjust the DirectRanker to text ranking, we extend the architecture of the networks $nn_1$ and $nn_2$. 
In contrast to the original paper, we included dropout layers and batch normalization~\cite{batchnorm} into the feature part to prevent the networks $nn_1$ and $nn_2$ from overfitting and to stabilize the learning process.
Some datasets provide the position of a focus word that indicates the most important word of the document in the corresponding ranking task.
For those datasets we add the additional networks $nn_{focus_1}$ and $nn_{focus_2}$ to the feature part that exclusively convert the focus words. 
The results of both the sentence network and focus word network are concatenated for the ranking part.
The ranking function is then given by:
\begin{equation}
    o_1 = \tau\left(w\left(\frac{(u_1, u_{f_1}) - (u_2, u_{f_2})}{2}\right)\right),
\end{equation}
where $u_{f_1} = nn_{focus_1}(x_1)$ and $u_{f_2} = nn_{focus_2}(x_2)$.
The corresponding loss function is determined as in Equation~\ref{equ:loss}.
In contrast to transforming sentence and focus word together in $nn_1$ and $nn_2$, we ensure by this procedure that the model is able to learn from the embeddings of two sources without the undesirable side-effect of one part receiving more weight than the other due to different feature distribution. 
\\
Instead of training the ranker only on the sparse pairwise labels, we calculate the BWS of the training set to create labels for all possible document pairs. 
The BWS is calculated for every document from pairwise comparisons as the fraction of times the document is labeled as relevant minus the fraction of times the document is labeled as less relevant.
During training we select the document pairs by random and set the label to 1 or -1 if the first document is more or less relevant according to the BWS. 
In contrast to the GPPL approach, this procedure encodes the annotation uncertainties in the BWS by assigning uncertain instances a low valued ranking.
Although the procedure depends on the BWS quality of the training set, we believe that the network benefits from being trained on more pairwise combinations especially if we reduce the amount of training data.
\subsection{Stacking}
\label{sec:stacking}
To improve the overall ranking performance, we combine the predictions of GPPL and the DirectRanker with stacking~\cite{WOLPERT1992241}. In stacking, a meta-model (also called the level-1 model) is learned to weight the predictions of the individual models (also called the level-0 models). The meta-model receives the output of the level-0 models and learns to which degree the models can be trusted. In order to prevent the meta-model from placing a higher weight on the ranking of a level-0 model that overfits on the training data, the level-0 models are trained on $n$ cross-validation splits. This procedure allows the meta-learner to evaluate the level-0 models on the unseen validation splits. In general, any classifier can be applied as a meta-model. \citet{Ting97stackedgeneralization} showed in an empirical study that simple linear models achieve the best performance as level-1 learners. Linear models offer the additional advantage that their learned parameters can easily be used to evaluate the output of level-0 models.

Through the use of stacking, $n$ meta-models and the corresponding level-0 rankers have to be trained. In order to benefit from their entire performance, we take the mean ranking of the $n$ meta-models. By this procedure not only the level-0 learners are combined, but also the individual meta-learners, whereby each one receives the same weight for the final ranking.

\subsection{Document Representation}
We investigate the combination with two document representations.
As a first approach we choose the~\emph{word2vec} embeddings trained on part of Google News~\cite{DBLP:journals/corr/MikolovSCCD13}.
We embed the words of the sentence and build the MWE to directly compare the findings of~\citet{simpson} with the DirectRanker.
However,~\emph{word2vec} embeddings have the disadvantage that
they assign only a single, fixed representation for each word, even though it may take on different meanings in different contexts, particularly with regard to creative language. 

To remedy these shortcomings, we introduce a second possibility, in which we apply the sentence transformers of~\citet{reimers-2019-sentence-bert} to generate meaningful sentence embeddings (SEs). In contrast to the MWEs, sentence transformers learn how to compose individual contextual word embeddings together to form SEs and are able to assign sentences with similar meaning a close representation in the vector space.
\section{Evaluation}\label{sec:evaluation}
\subsection{Datasets}
We explore the introduced models on two datasets including different types of creative language. First, the humor dataset~\cite{simpson} that contains 4030 samples with various degrees of humorousness. The humour dataset is an extension of~\citet{miller}, which includes 3398 humorous and 632 non-humorous samples with an average sentence length of 11 words. The humor section can be grouped into homographic and heterographic puns containing purely verbal humour while the non-humorous section contains proverbs and aphorisms.

Second, the metaphor dataset~\cite{weeding}, which contains 72816 samples annotated for metaphor novelty.
The dataset is a labeled version of the VU Amsterdam Metaphor Corpus~\cite{22089106df104bd3bd84d5e686d5bed2} that tags metaphors in four genres: news, fiction, conversation transcripts, and academic texts.

Both datasets were labeled using crowdsourcing. On the humor dataset every instance was selected for 14 random pairwise comparisons and each pair was labeled by 5 different annotators. On the metaphor dataset each instance was selected for 6 random tuples containing 4 instances, with each tuple labeled by 3 annotators. To generate pairwise comparisons, we adopt the procedure of~\citet{simpson}, considering only the most novel and most conventionalized sample of the tuple. The resulting pairwise comparisons are labeled 1.55 times on average and each sample is present in 8.6 pairs on average.

\subsection{Experimental Setting}
\begin{table*}[t]
    \centering
    \caption{Mean results with different training data sizes on the humour and metaphor dataset. We show the Spearman's $\rho$ against the BWS. The $\diameter$ indicates that the model's mean ranking of the 4-fold cross validation ensemble is evaluated (see Section \ref{sec:stacking} for more details). In stacking we first name the embedding of GPPL and then the embedding of the DirectRanker.}
    \begin{tabular}{lcccc|cccc}
    &\multicolumn{4}{c|}{Humour}&\multicolumn{4}{c}{Metaphor}\\
         &60\%&33\%&20\%&10\%&60\%&33\%&20\%&10\%\\
         \hline
         \hline
         Bert Baseline & \textbf{0.62} & 0.44 & 0.20 & 0.12 & 0.38 & 0.35 & 0.28 & 0.20 \\
         Bert + Focus Word & \multicolumn{4}{c|}{-}& 0.53 & 0.47 & 0.39 & -0.03 \\ 
         \hline
         GPPL MWE&0.54&0.53&0.47&0.41&0.58&0.55&0.51&0.35\\
         DirectRanker MWE&0.54&0.50&0.44&0.30&0.64&0.60&0.52&0.37\\
         \hline
         Stacking SE/SE&0.61&0.59&0.53&0.43&\textbf{0.69}&0.64&0.56&0.39\\
         Stacking SE/MWE&0.61&\textbf{0.60}&\textbf{0.56}&\textbf{0.46}&\textbf{0.69}&0.63&0.58&0.43\\
         Stacking MWE/MWE&0.58&0.56&0.51&0.41&0.68&0.64&0.61&0.53\\
         GPPL $\diameter$ MWE&0.58&0.55&0.47&0.43&0.59&0.55&0.48&0.29\\
         GPPL $\diameter$ SE&0.59&0.58&0.53&\textbf{0.46}&0.62&0.56&0.48&0.34\\
         DirectRanker $\diameter$ MWE&0.55&0.52&0.46&0.35&0.67&0.63&0.60&0.53\\
         DirectRanker $\diameter$ SE&0.60&0.57&0.51&0.42&0.68&0.63&0.58&0.47\\
        \hline
         Stacking Focus Word&\multicolumn{4}{c|}{-}&0.68&\textbf{0.65}&\textbf{0.62}&\textbf{0.57}\\
         GPPL $\diameter$ Focus Word&\multicolumn{4}{c|}{-}&0.60&0.56&0.48&0.40\\
         DirectRanker $\diameter$ Focus Word&\multicolumn{4}{c|}{-}&0.68&\textbf{0.65}&\textbf{0.62}&\textbf{0.57}\\
   
         \hline

    \end{tabular}
    \label{tab:humour}
\end{table*}
We evaluate our experiments using 3 random training and test splits and show the average results. To examine the ranking performance on sparse data, we also experiment with artificially reducing training set sizes.  For this purpose, we randomly select 60\%, 33\%, 20\%, and 10\% of the document IDs and use only those pairwise annotated samples for training that include both IDs in our selection. The remaining samples are used in the test set to counteract the model variation for smaller training set sizes.

To generate the SEs we use the pretrained 'bert-base-nli-stsb-mean-tokens' model for both datasets.
While we obtain the training SEs only from the corpus of training data, we combine the corpus of training and test data to generate the test embeddings. Although the test embeddings would be better adapted to the training data if each test document was encoded individually with the training corpus, this would lead to an increased calculation overhead by the continuously repeated execution of the sentence transformer for each document. However, our experiments show that even with our test setting, satisfactory results can still be achieved.

The configurations of the individual methods are described in the following.
For the DirectRanker we set the dimensions of the 4 MLP feature part to 2k, 500, 64, and 7.
Furthermore, we use Adam as the optimizer with a learning rate of 0.001, a dropout rate of 0.4 and $\tanh$ as activation function for both model parts in the DirectRanker.
The GPPL is trained with the same configurations as by~\citet{simpson}, including a Mat{'e}rn $\frac{3}{2}$ kernel and 500 inducing points. In addition to the MWEs, they investigate the combination of different linguistic features. We include their best combination in the GPPL: the average token frequency (taken from a 2017 Wikipedia dump) and the average bigram frequency (taken from Google Books Ngrams).
Stacking is performed on 4 cross-validation splits and we choose a linear regressor as level-1 model.

\subsection{Results}
\label{sec:results}
\begin{figure*}[t]
\centering
   \includegraphics[width=1.0\textwidth]{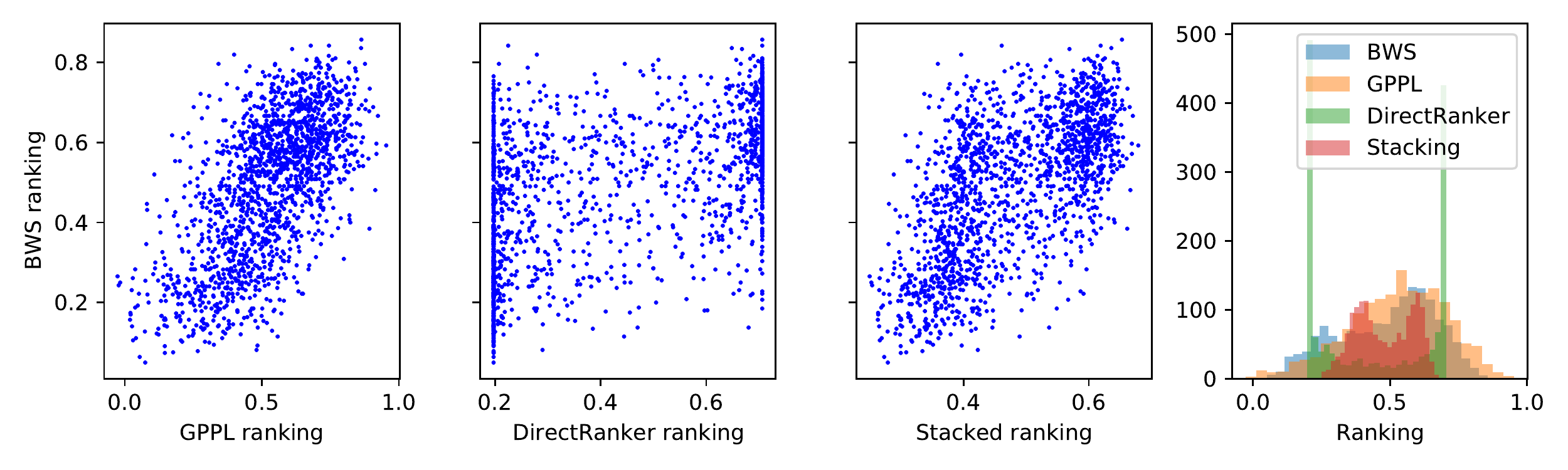}
\caption{Comparison of the ranking distribution on the humour data in the 60\% case with SEs for GPPL and MWEs for the DirectRanker.
The first three plots show the linear correlation between the BWS gold standard and the results of the different methods. The last plot summarizes the corresponding ranking distributions.}
   \label{fig:humour_ranking} 
\end{figure*}
To illustrate key differences between the models using the two embeddings, we provide the results of the combinations with the most impact.
The linear rank correlation between the prediction and the BWS gold standard is obtained with Spearman's $\rho$ and used as quality measure.
In order to distinguish the influence of the different methods, the results of the individual models, stacking, and the intermediate ensemble rankers are shown in Table~\ref{tab:humour}. As a additional baseline, we include pretrained BERT regression models (bert-base-cased) fine-tuned on the respective training sets. For the metaphor data, we also include a focus word model, which incorporates a fine-tuned BERT model along with the \emph{word2vec} embedding of the focus word. 
In a second step GPPL and the DirectRanker trained on the complete training splits are evaluated. Finally, we compare the results with the mean ranking of the model ensemble learned during the 4 fold cross-validation. The third row of the tables provides the results of~\citet{simpson} adapted to our evaluation setting. The standard deviation of the individual methods ranges from 0.016 for 60\% to 0.038 for 10\% on the humour dataset, and from 0.006 for 60\% to 0.043 for 10\% on the metaphor dataset.

On the humour data set the BERT baseline performs best in the 60\% case as it is able to classify the less relevant documents better. However, the baseline is not well suited in other scenarios with less data in which the investigated pairwise models achieve significantly better results.
GPPL outperforms the DirectRanker on almost all training set sizes and document representations. Both the GPPL and the DirectRanker benefit from the SEs, whereas the DirectRanker is able to achieve an increased benefit especially for small training sizes.
Generally, the DirectRanker does not generalize well with the MWEs in the case of 10\% training set.
By combining the weighted ranking of the GPPL on the SEs and the DirectRanker on MWEs, stacking is able to improve the individual performances for the 60\% - 20\% training sets. Only in the 10\% case stacking is just able to balance the high performance difference of the two level-0 models.
We believe that the combination of representations with different information content between the level-0 models may help to compensate their systematic errors and improve the stacking performance.
If both level-0 models use the same representation as for example the SEs, stacking balances the performance of both models but is not able to amplify the performance much compared to the individual rankers.

On the metaphor dataset the behavior of the investigated individual models changes in some cases. The BERT baseline is not able to reach competitive results in all training scenarios. While the DirectRanker also benefits from the mean ranking of the cross-validation models, the performance of the GPPL ensemble decreases for smaller training sets. We assume that the high performance gain of the DirectRanker's cross-validation models results from the integration of the validation set to choose the best model. Without early stopping according to the validation set, the neural network tends to increased overfitting on this dataset. The DirectRanker outperforms GPPL in almost all test cases on the same document representation. 
Stacking balances the ranking performance of the level-0 models in the most settings. 
In the 20\% and 10\% case stacking decreases the maximum individual performance on the SEs as the GPPL overfits on the validation set. This might be an effect of the SE construction. While training and validation set are transformed in one corpus, the test set is transformed with the training and validation set, making the SEs of the validation split more comparable to the training set than to the test set.
By combining models trained on different representations, stacking is more likely to balance the individual predictions.

For metaphor novelty,
the results of the models trained on only the focus word embedding 
or only the sentence representation 
indicate a heavy reliance on the focus word. While neither GPPL nor DirectRanker is able to extract any useful information from the sentences alone (with SEs in the 60\% case DirectRanker and GPPL reach a Spearman's $\rho$ of 0.17 and 0.4, respectively), taking only the focus word under consideration instead of both information sources has a beneficial effect on the ranking performance in the 33\% - 10\% case.
We assume that the label decisions were influenced by the partial exclusive consideration of the focus word during the annotation process. Further analyses in Section~\ref{sec:analysis} support this assumption.
\subsection{Analysis}
\label{sec:analysis}
In order to work out the differences between the presented models, we first take a closer look at the ranking distribution on the test set (rankings $r$ are all shifted by $(r+1)*0.5$). Figure~\ref{fig:humour_ranking} summarizes the results on the humour data in the 60\% case with SEs for GPPL and MWEs for the DirectRanker (the rankings on the metaphor dataset are similar distributed and not shown). We plotted the individual rankings against the BWS gold standard to visualize the linear correlation with the investigated methods and show the individual ranking frequencies in a histogram. While the GPPL scores tend to a Gaussian distribution with a mean around 0.6, the DirectRanker predicts more extreme values around 0.2 and 1 as we use $\tanh$ as activation function. By combining the methods with stacking, the ranking scores acquire the characteristics of both methods. 
\begin{figure*}[t]
    \centering
   \includegraphics[width=1.0\textwidth]{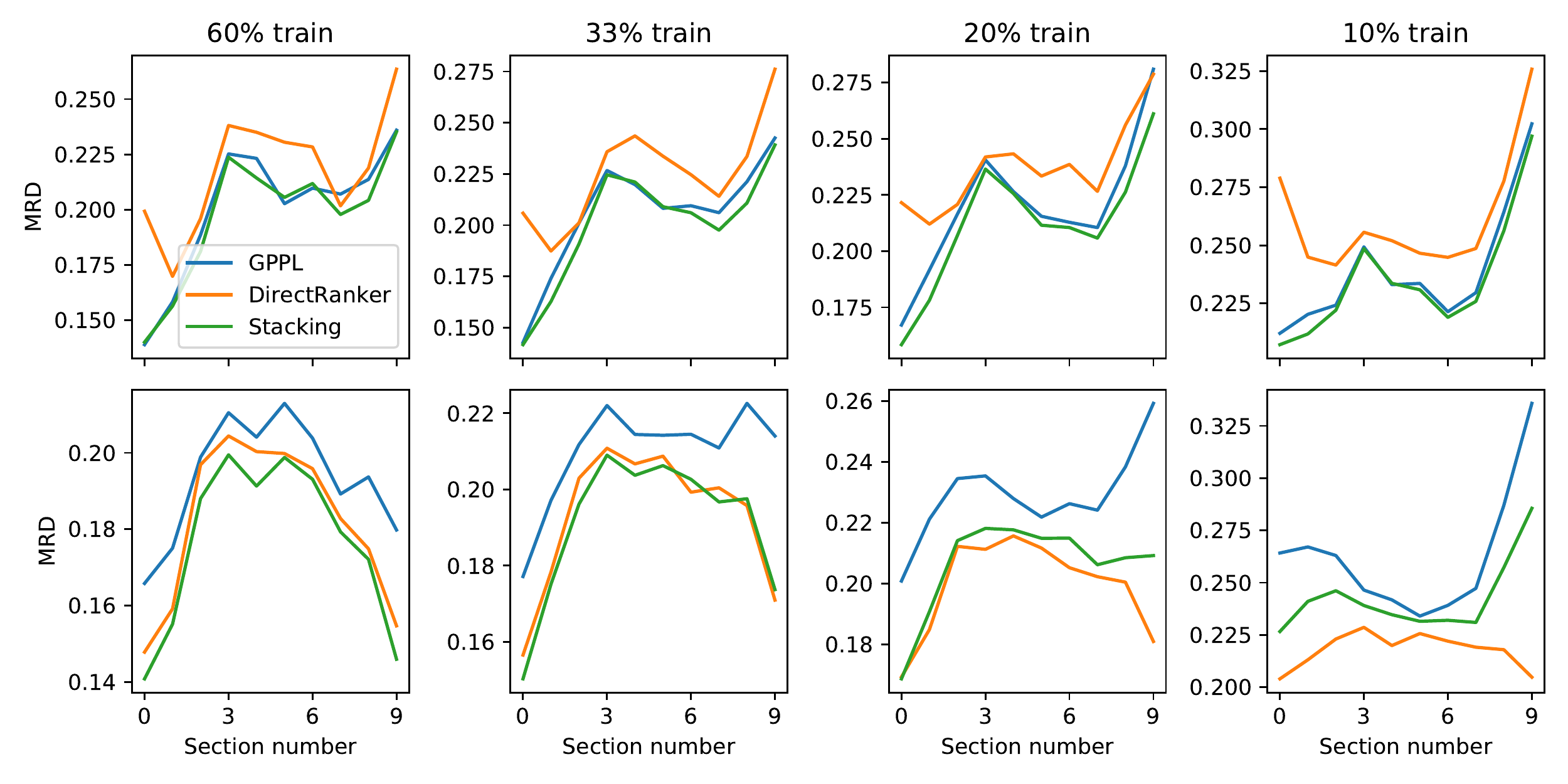}
\caption{Mean ranking distance with SEs for GPPL and MWEs for the DirectRanker. The upper plots show the performance on the humour dataset and the lower plots on the metaphor dataset for all training percentages. Instances are grouped in segments of equal size according to the BWS score whereas the segment number corresponds to the ranking level. The MRD shows the mean ranking distance of the entire ranking in each segment.}
   \label{fig:ranking_sections}
\end{figure*}

Additionally, we plot the ranking quality for all training percentages in Figure~\ref{fig:ranking_sections} on 10 equally sized distinct segments of the test set divided according to the gold score. The ranking quality is evaluated with the mean ranking distance (MRD) by subtracting the global gold document position from the predicted global document position and dividing the result by the number of all test documents times the bin size. Despite the differences in ranking frequencies, the performance of the methods within the different sections is similar in the most cases. On the humour dataset the GPPL performs better on lower ranked documents which are mostly represented by proverbs and aphorisms on the fist 3 segments. With less training data the DirectRanker performance decrease on these segments. However, combining the DirectRanker with SEs leads to a performance similar to GPPL (not shown in the plots).
After segment 7 the ranking performance of all models decreases on more relevant instances. While the performance difference between section 0 and 9 stays around 0.1 on all train sizes, the difference between section 0 and 3-7 decreases for smaller train sizes. The behavior on the metaphor dataset is similar with the exception that the models perform on the relevant instances almost as well as on the lower ranked instances. These documents correspond to the metaphors with more ordinary focus word like 'take', 'make', and 'get' (with low metaphorical novelty) and metaphors with infrequent focus word like 'bloody', 'rapprochement', and 'explode'. This observation also supports the assumption that some of the instances were only annotated regarding the focus word. Only in the 20\% and 10\% cases the GPPL and the stacking (recall Section~\ref{sec:results}) deteriorate on the sections 6-9.


\section{Conclusion}\label{sec:conclusion}
In this work we investigated a pairwise ranking approach for creative language based on adapting a recent neural architecture, DirectRanker, that can learn efficiently from small training sets.
We combined it with a Bayesian model, GPPL, using stacking and evaluated the behavior of all models on the tasks of predicting the humorousness and metaphorical novelty with different document representations.
During our experiments the models differ in their ranking performance according to the specific ranking task. A closer look reveals that the ranking quality generally differs in three sections of the low, high and medium ranked documents. By using less training samples, the performance of low and medium ranked documents becomes more comparable while more relevant documents are more difficult to categorize.
Despite the expectation that neural networks suffer from overfitting on small data sets, the DirectRanker was able keep up with the performance of the GPPL or even improved on it,
while a standard BERT-based approach performed poorly. We showed that the GPPL clearly benefits from sentence embeddings while this statement cannot be absolutely confirmed for the DirectRanker. However, stacking was able to balance the individual predictions and improves them especially if the underlying level-0 models were trained on diverse document representations. Therefore, we would suggest to combine sentence embeddings with the GPPL and mean word embeddings with the DirectRanker for stacking. 
The proposed stacking approach clearly outperforms state-of-the-art results and is also applicable in the case of limited data.
\bibliography{anthology,eacl2021}
\bibliographystyle{acl_natbib}
\end{document}